\documentclass[10pt,twocolumn,letterpaper]{article}

\usepackage{cvpr}              
\definecolor{cvprblue}{rgb}{0.21,0.49,0.74}
\usepackage[pagebackref,breaklinks,colorlinks,allcolors=cvprblue]{hyperref}

\usepackage{booktabs}
\usepackage{multirow}
\usepackage{makecell}
\usepackage{colortbl}
\usepackage{fontawesome5}

\def\paperID{18266} 
\def\confName{CVPR}
\def\confYear{2026}

\title{Mantis: A Versatile Vision-Language-Action Model \\ with Disentangled Visual Foresight}

\author{
Yi Yang\textsuperscript{1,2}\;\;\; 
Xueqi Li\textsuperscript{2,3}\;\;\; 
Yiyang Chen\textsuperscript{1}\;\;\; 
Jin Song\textsuperscript{4}\;\;\; 
Yihan Wang\textsuperscript{5} 
\\
Zipeng Xiao\textsuperscript{1}\;\;\; 
Jiadi Su\textsuperscript{5} \;\;\; 
You Qiaoben\textsuperscript{6} \;\;\; 
Pengfei Liu\textsuperscript{1,2} \;\;\;
Zhijie Deng\textsuperscript{1,\thanks{Corresponding author}} 
\vspace{2mm}
\\
\textsuperscript{1}$\texttt{SJTU}$\;\;\;
\textsuperscript{2}$\texttt{SII}$\;\;\;
\textsuperscript{3}$\texttt{SUSTech}$\;\;\;
\textsuperscript{4}$\texttt{NJUPT}$\;\;\;
\textsuperscript{5}$\texttt{FDU}$\;\;\;
\textsuperscript{6}$\texttt{BOSCH}$
\vspace{1mm}
\\
{\small \faGithub\ Code: \texttt{\url{https://github.com/SJTU-DENG-Lab/Mantis}}}
}

\begin{document}

\maketitle

\begin{abstract}


Recent advances in Vision-Language-Action (VLA) models demonstrate that visual signals can effectively complement sparse action supervisions. However, letting VLA directly predict high-dimensional visual states can distribute model capacity and incur prohibitive training cost, while compressing visual states into more compact supervisory signals inevitably incurs information bottlenecks. Moreover, existing methods often suffer from poor comprehension and reasoning capabilities due to the neglect of language supervision. This paper introduces Mantis, a novel framework featuring a Disentangled Visual Foresight (DVF) to tackle these issues. Specifically, Mantis decouples visual foresight prediction from the backbone with the combination of meta queries and a diffusion Transformer (DiT) head. With the current visual state provided to the DiT via a residual connection, a simple next-state prediction objective enables the meta queries to automatically capture the latent actions that delineate the visual trajectory, and hence boost the learning of explicit actions. The disentanglement reduces the burden of the VLA backbone, enabling it to maintain comprehension and reasoning capabilities through language supervision. Empirically, pretrained on human manipulation videos, robot demonstrations, and image-text pairs, Mantis achieves a 96.7\% success rate on LIBERO benchmark after fine-tuning, surpassing powerful baselines while exhibiting high convergence speed. Real-world evaluations show that Mantis outperforms $\pi_{0.5}$, a leading open-source VLA model, particularly in instruction-following capability, generalization to unseen instructions, and reasoning ability. We also introduce the adaptive temporal ensemble (ATE) strategy to balance computational efficiency and motion stability during inference, yielding the Mantis-ATE variant, which reduces inference counts by 50\% while maintaining performance. Code and weights are released to support the open-source community.

\end{abstract}    
\section{Introduction}
\label{sec:intro}
Robotic learning has witnessed remarkable progress, particularly in developing robust control strategies for diverse tasks across heterogeneous environments. 
Among the most promising approaches are Vision-Language-Action (VLA) models~\cite{kim2024openvla, brohan2022rt, mu2023embodiedgpt, zawalski2024robotic, zitkovich2023rt, cheang2024gr, team2024octo}, which leverage pretrained Vision-Language Models (VLMs)~\cite{beyer2024paligemma, bai2025qwen2} to translate linguistic instructions and visual observations into executable robotic actions. 
Despite these advancements, existing VLA approaches face a fundamental challenge: the low-dimensional action signals can be too sparse to adequately supervise the large VLA model that processes high-dimensional sensory inputs~\cite{li2025drivevla}. 
This mismatch leaves much of the model’s representational capacity underutilized, thereby constraining overall performance.

A remedy is to integrate the visual foresight prediction problem into VLA training, with the model asked to predict dense future visual states in addition to forecasting actions~\cite{zhao2025cot, wang2025unified}. 
However, high-dimensional visual states usually contain redundancies that distract the model from action prediction, resulting in high training costs and slow convergence during downstream fine-tuning~\cite{gao2025vla}. 
Alternative approaches compress visual states into more compact supervisory signals~\cite{wen2023any, ye2024latent}. 
Yet, compression inevitably diminishes subtle variations among visual states that convey fine-grained motions, thereby creating an information bottleneck. 
Furthermore, existing methods often overlook language supervision, rendering the context understanding and reasoning capabilities of the model unguarded~\cite{zhou2025chatvla}.

We propose Mantis, a novel VLA model featuring a Disentangled Visual Foresight (DVF) to address these issues.
As illustrated in Figure~\ref{fig:arch}, Mantis combines meta queries~\cite{pan2025transfer} with a Diffusion Transformer (DiT)~\cite{peebles2023scalable} head to predict future visual states.
Despite its simplicity, this design can effectively disentangle the tight coupling of action learning and foresight prediction, considering its efficacy in text-to-image generation tasks~\cite{pan2025transfer, chen2025blip3}.
Furthermore, we advocate feeding the current visual state also to the DiT through a residual connection~\cite{he2016deep}, which enables the meta queries to automatically capture the inter-frame dynamics that delineate the visual trajectory. 
Intuitively, these queries extract latent actions which can facilitate more effective explicit action generation; we therefore term them \emph{latent-action queries}. 
Then, we use learnable \emph{action queries} to extract information from both the input and latent-action queries via causal attention.
We apply an action head to map the extraction outcomes to explicit actions as in \cite{zhang2025dreamvla}.


The disentangled design of Mantis reduces the representation burden on the VLA backbone, thereby freeing up more capacity for language supervision, which helps preserve the model’s semantic understanding and reasoning capabilities after action learning.
To minimize the competition between learning signals from various modalities, we develop a progressive training recipe that incrementally incorporates additional modalities for smooth fusion. 
For inference, Mantis leverages the Temporal Ensemble~\cite{zhao2023learning} method for motion stability.
Considering its inefficiency issues, we also propose Adaptive Temporal Ensemble (ATE), which dynamically adjusts the ensemble strength according to the motion stability demand of the current timestep. \Eg, fine-grained object manipulation can require much higher motion stability than unladen movements. 
We term the corresponding model variant as Mantis-ATE.

We pretrain Mantis based on three data sources: the SSV2 dataset~\cite{goyal2017something} containing 220K human manipulation videos, the DROID dataset~\cite{khazatsky2024droid} with 76K robot demonstrations (covering video and action), and image–text pairs drawn from 38 multimodal datasets. 
We then evaluate Mantis on the LIBERO simulation benchmark~\cite{liu2023libero} and in real-world settings. Extensive experiments validate our approach from multiple perspectives:
(1) Mantis surpasses several strong baselines~\cite{wen2023any, zhao2025cot, cen2025worldvla, bu2025learning, wang2025unified, lv2025f1}, achieving 96.7\% success on the LIBERO benchmark and showing faster convergence than previous visual foresight approaches~\cite{wang2025unified}.
(2) Experiments on the Agilex platform confirm Mantis’s robust instruction-following and generalization capabilities. Across three test scenarios, it accurately executes in-domain commands and effectively generalizes to out-of-domain instructions, outperforming the leading open-source VLA model $\pi_{0.5}$~\cite{intelligence2025pi_}.
(3) The Mantis-ATE variant reduces inference calls by up to 50\% while maintaining comparable task success rates.

In summary, our main contributions are three-folds:
\begin{itemize}
\item We propose Disentangled Visual Foresight (DVF) that provides concise and instructive look-ahead cues for action prediction and construct a new VLA model Mantis.
\item We design a progressive training recipe for modality fusion, enabling Mantis to effectively integrate action prediction with language understanding.
\item Mantis achieves 96.7\% success on the LIBERO benchmark and demonstrates superior instruction-following capabilities in real-world robot experiments.
\end{itemize}

\section{Related Work}
\label{sec:related_work}

\subsection{Vision-Language-Action Models}
The rapid evolution of vision-language models (VLMs)~\cite{liu2023visual, bai2025qwen2, beyer2024paligemma, team2024chameleon, wu2024vila} has catalyzed the emergence of Vision-Language-Action (VLA) models~\cite{zitkovich2023rt, kim2024openvla, brohan2022rt, mu2023embodiedgpt, zawalski2024robotic, cheang2024gr, team2024octo, tian2024predictive, wang2025vla, wen2025tinyvla, li2023vision}.
These systems integrate a vision-language backbone with an action prediction component, extending their functional capabilities beyond pure perception.
By leveraging the rich perceptual and linguistic representations of pretrained VLMs, VLA models enable robots to interpret and execute human commands with improved adaptability, transcending the limitations of purely reactive control policies.
Despite this potential, existing VLA models suffer from limited understanding capabilities. Robot-specific training often overwrites the critical vision-text alignments acquired during pretraining~\cite{zhou2025chatvla}, degrading instruction-following performance and undermining fundamental reasoning abilities.
While some approaches employ explicit language assistance to mitigate this issue~\cite{zawalski2024robotic, sun2025emma}, they are prone to error accumulation~\cite{gao2025vla}. Other methods introduce implicit language supervision through auxiliary losses on language representations~\cite{zhou2025chatvla, ji2025robobrain, chen2025training}, but lack the visual foresight capabilities necessary for frame-level planning.
In contrast, we propose a framework that incorporates visual foresight while simultaneously preserving semantic understanding and reasoning capabilities.

\subsection{Vision-Augmented Action Learning}
Vision-augmented action learning effectively complements sparse action signals and better exploits the representational capacity of VLA models. We categorize vision-augmented action learning approaches as follows:

\begin{figure}[htbp]
    \centering
    \includegraphics[width=0.85\linewidth]{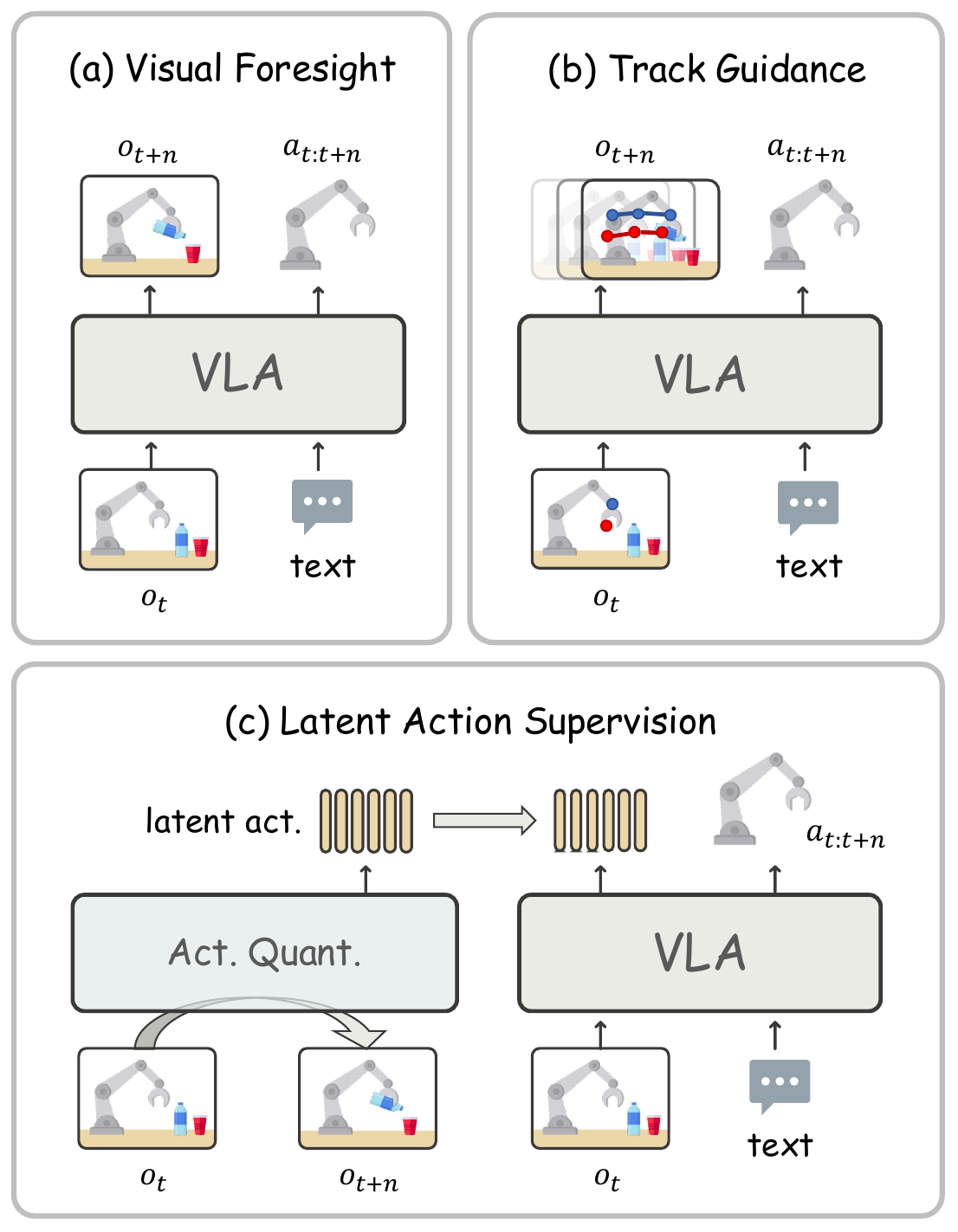}
    \caption{\textbf{Vision-augmented action learning paradigms.} (a) Visual Foresight enhances action prediction by forecasting future frames. (b) Track Guidance employs compressed visual state representations to guide action prediction. (c) Latent Action Supervision improves action learning through auxiliary latent actions.}
    \label{fig:vision}
\end{figure}

\textbf{1) Visual Foresight.} This approach enhances action prediction by forecasting future frames, using either explicit or implicit methods.
Explicit methods generate future frames through autoregressive discrete image tokens~\cite{zhao2025cot,lv2025f1} or learnable query tokens with a vision decoder~\cite{wu2023unleashing,zhang2025dreamvla}, which then inform action prediction.
Implicit methods jointly train the VLA model on video generation and action prediction tasks, establishing an implicit connection between visual forecasting and action prediction~\cite{cen2025worldvla,wang2025unified,jiang2025rynnvla}. 
However, pixel-level forecasting can introduce unnecessary information, distracting the model from action prediction and resulting in high training costs.
Moreover, this approach may lead the model to mistakenly associate physical motion with visual appearances, such as texture or lighting changes, potentially resulting in hallucinations~\cite{du2023learning}.

\textbf{2) Track Guidance.} 
To mitigate these limitations, several studies compress visual states into compact, control-oriented representations such as keypoint tracks~\cite{wen2023any,bharadhwaj2024track2act,bharadhwaj2024gen2act}.
These tracks capture essential physical dynamics and facilitate action prediction.
Nonetheless, this compression can cause information bottlenecks, and the precision of point-tracking extracted from videos is often limited, leading to higher action prediction errors.

\textbf{3) Latent Action Supervision.}
Other approaches employ latent actions to supervise action prediction~\cite{ye2024latent,bu2025learning,chen2025moto}.
Typically, an action quantization model is first trained to learn discrete latent actions from inter-frame differences.
The VLA model is then trained to predict these latent actions before being fine-tuned on robot manipulation data.
This strategy leverages the insight that inter-frame dynamics can be represented as action primitives that aid prediction.
However, training an additional quantization model increases computational complexity.
An overview of these vision-augmentation paradigms is provided in Fig.~\ref{fig:vision}.
In contrast to prior work, Mantis decouples visual foresight prediction from the backbone, yielding more compact and accurate auxiliary information for action prediction.

\section{Methodology}
\label{sec:methodology}

\begin{figure*}[t]
    \centering
    \includegraphics[width=1.0\linewidth]{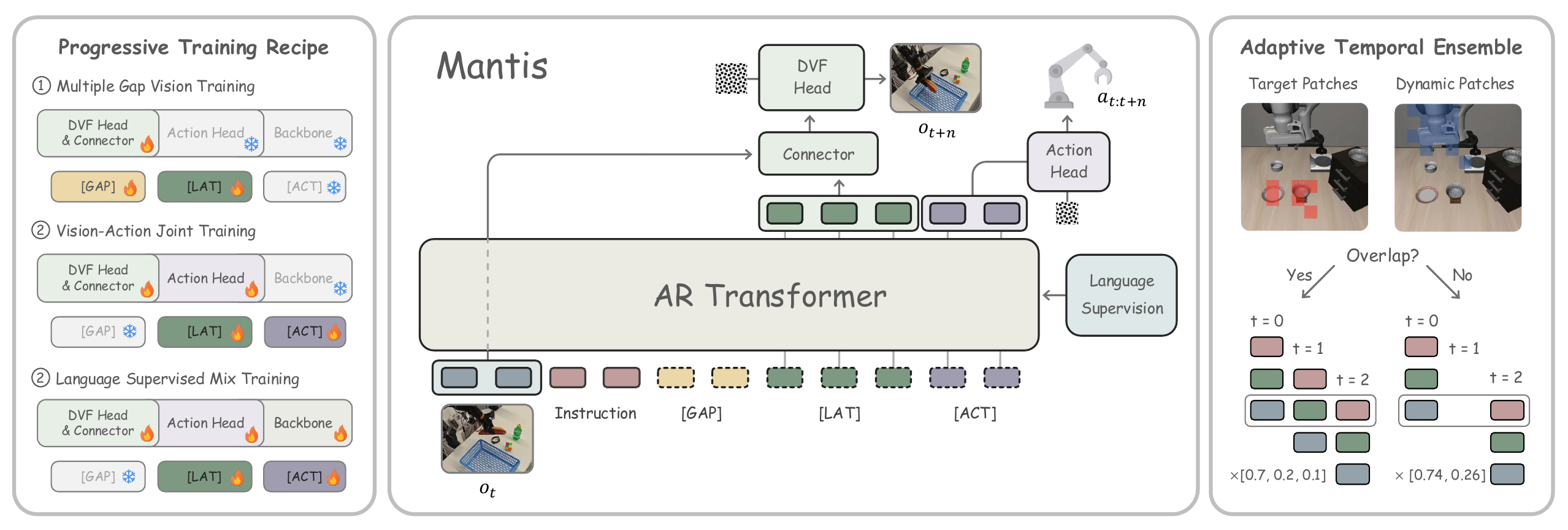}
    \caption{\textit{Left:} \textbf{Progressive training recipe.} Mantis progressively integrates multiple modalities to achieve stable and well-balanced optimization. \textit{Center:} \textbf{Overview of Mantis.} The framework consists of a backbone network, a DVF head, and an action head. The DVF head predicts future frames to facilitate latent action learning, thereby improving action prediction. Language supervision helps maintain the backbone’s capability for understanding and reasoning. \textit{Right:} \textbf{Adaptive Temporal Ensemble.} Mantis-ATE dynamically adjusts the ensemble strength based on the overlap between target tokens and dynamic tokens.}
    \label{fig:arch}
\end{figure*}

This section provides an overview of Mantis, beginning with its model architecture and specifications, followed by detailed descriptions of the progressive training recipe and the adaptive temporal ensemble mechanism.

\subsection{Model Overview}

As illustrated in Fig.~\ref{fig:arch}, Mantis comprises the following components: a backbone model $\mathcal{P}$, a connector $\mathcal{C}$, a DVF head $\mathcal{D}$, an action head $\pi$, a set of trainable latent-action queries $\texttt{[LAT]}$, and action queries $\texttt{[ACT]}$.
The task is specified by a language instruction $l$.
At each time step $t$, the backbone $\mathcal{P}$ receives $l$ and a visual state (\eg, a raw image frame) $\mathbf{o}_t$. 
These inputs, together with $\texttt{[LAT]}$, are packed into a sequence and mapped as 
\begin{equation}
  \mathbf{h}_t=\mathcal{P}(\mathbf{o}_t, l, \texttt{[LAT]}).
  \label{eq:P_theta}
\end{equation}
$\mathbf{h}_t$ are then concatenated with $\mathbf{o}_t$ and fed into the connector $\mathcal{C}$, which projects them into the conditional input of $\mathcal{D}$ to generate the future image frame $\mathbf{o}_{t+n}$ with $n$ as the time step gap between the current and future frames:
\begin{equation}
  \mathbf{o}_{t+n}=\mathcal{D} (\mathcal{C}(\mathbf{o}_{t}, \mathbf{h}_{t})).
  \label{eq:o_t_n}
\end{equation}

This design prevents the VLA backbone from producing redundant visual information directly. 
Besides, feeding $\mathbf{o}_t$ into $\mathcal{D}$ via a residual connection~\cite{he2016deep} enables the $\texttt{[LAT]}$ queries to capture inter-frame dynamics that characterize the visual trajectory rather than reconstructing complete frames. 
These dynamics connect to \emph{latent actions}—a visual manifestation of explicit robot motions—thereby providing targeted guidance for action prediction. 

Next, 
the actions for the next $n$ time steps can be generated by the action head $\pi$ via
\begin{equation}
  \mathbf{a}_{t:t+n}=\pi (\mathcal{P}(\mathbf{o}_{t}, l, \texttt{[LAT]}, \texttt{[ACT]})),
  \label{eq:a_t_a_t_n}
\end{equation}
where the latent-action queries $\texttt{[LAT]}$ is added to the context and the action queries $\texttt{[ACT]}$ are used to extract information from the context. 

Furthermore, to produce denser visual predictions during training and accommodate diverse downstream tasks, we introduce multi-gap queries \texttt{[GAP]}. These queries are inserted before \texttt{[LAT]} to guide the generation of future frames at varying time step intervals.




\subsection{Model Specification}



\noindent \textbf{VLM Backbone.} 
Considering the robust understanding and reasoning capabilities of Qwen2.5-VL~\cite{bai2025qwen2}, we adopt it as our backbone.
Qwen2.5-VL natively supports flexible input resolutions, allowing us to assign higher resolution to the primary camera while allocating lower resolution to the wrist camera, which captures less spatial detail.
\vspace{0.5em}

\noindent \textbf{DVF Head.}
We employ Sana~\cite{xie2024sana} as the DVF head due to its superior performance in text-to-image generation. Sana is a highly efficient DiT that integrates a deep compression autoencoder~\cite{chen2024deep}.
We define a connector comprising 12 transformer encoder layers and a projection layer that bridges the backbone's output space with the DiT's input space, following the Qwen2.5 LLM architecture~\cite{yang2025qwen3} while using bidirectional attention.
During action inference, the DVF head is omitted to reduce computational overhead, as visual state prediction is not required for robot execution.

\vspace{0.5em}
\noindent \textbf{Action Head.} 
Following~\cite{zhang2025dreamvla}, we employ a DiT-based action head for action prediction. Learnable action queries first aggregate information from input and latent-action queries via causal attention, then the action head is used to denoise Gaussian noise into an $n$-step action trajectory.

\subsection{Progressive Training Recipe} 
Directly fusing vision, language, and action during pretraining can bias learning toward the easiest signal (\eg, actions) or overfit dominant modalities (\eg, language), causing cross-modal competition and unstable convergence.
To address this, we adopt a progressive training recipe that introduces modalities in stages, promoting more stable optimization (Fig.~\ref{fig:arch}, left). The training proceeds in three stages:


\vspace{0.5em}
\noindent \textbf{Stage 1: Multiple Gap Vision Training.}
We first train Mantis on videos without action annotations to predict future frames, encouraging the model to infer latent actions from visual dynamics. We use human manipulation videos so that Mantis learns general manipulation skills and broad world knowledge~\cite{ye2024latent}. 
\begin{figure}[t]
    \centering
    \includegraphics[width=1.0\linewidth]{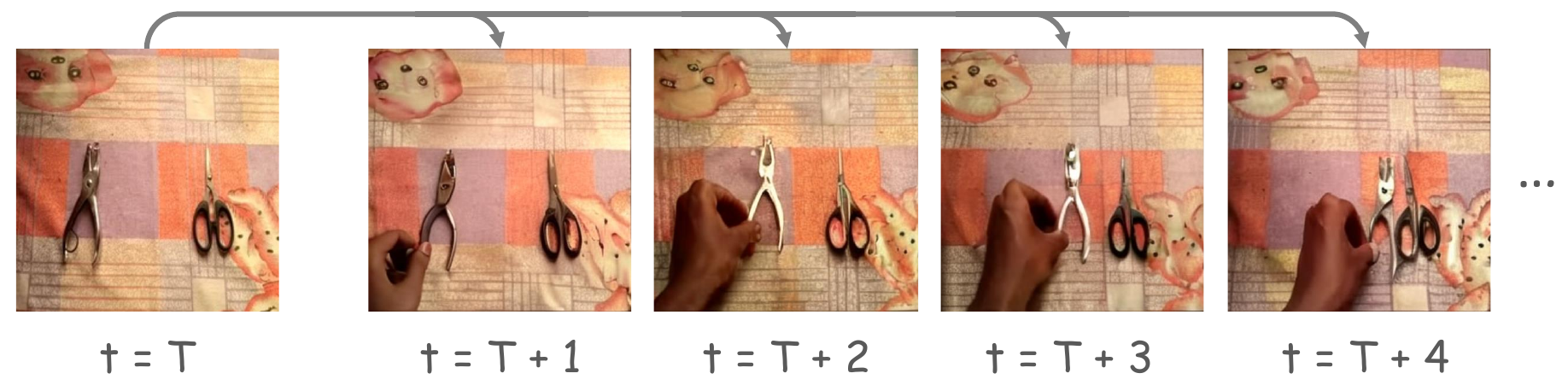}
    \caption{\textbf{Visualization of multi-gap future frame generation.}}
    \label{fig:multi}
\end{figure}
During training, we unfreeze the DVF head and latent-action queries, optimizing the diffusion loss $\mathcal{L}_{\text{DVF}}$ while keeping the backbone frozen to preserve its pretrained language representations. Additionally, the multi-gap queries are unfrozen to guide future frame generation across different time step intervals. Fig.~\ref{fig:multi} visualizes examples of multi-gap future frame generation.

\vspace{0.5em}
\noindent \textbf{Stage 2: Vision-Action Joint Training.}
Then we introduce the action modality using a robot demonstration dataset. The time step gap is fixed to match the action chunk size for temporal alignment between visual and action streams. The model is optimized with $\alpha \mathcal{L}_\text{DVF} + \mathcal{L}_\text{action}$, where $\mathcal{L}_\text{action}$ denotes the action head’s diffusion loss and $\alpha$ balances the terms. In this stage, we unfreeze the action queries while keeping the backbone frozen.

\vspace{0.5em}
\noindent \textbf{Stage 3: Language Supervised Mix Training.}
To supervise the language modality, we jointly train on a collection of multimodal datasets together with the robot demonstration data. 
The backbone is unfrozen, and a cross-entropy loss $\mathcal{L}_{\text{lang}}$ is applied to the language outputs.
The overall objective is $\alpha \mathcal{L}_\text{DVF} + \mathcal{L}_\text{action} + \beta \mathcal{L}_\text{lang}$, where $\alpha$ and $\beta$ weight the contributions of the vision and language losses. This progressive fusion yields stable and efficient optimization, producing a vision-augmented foundation with robust multimodal understanding for downstream tasks.

\subsection{Adaptive Temporal Ensemble} 
During inference, Mantis incorporates the commonly used Temporal Ensemble method~\cite{zhao2023learning} to enhance motion stability.
Considering its high computational overhead, we also introduce an Adaptive Temporal Ensemble (ATE) strategy that dynamically adjusts the ensemble strength according to the motion stability required at each inference time step. 

As shown in Fig.~\ref{fig:arch} (right), ATE maintains two sets of input vision patches at each time step.
\textbf{(1) Target patches} denote image regions that are most relevant to the language instruction. Following~\cite{xu2025vla}, we compute text-to-vision attention scores from Mantis backbone’s cross-attention module and select the top $\tau_\text{target}\%$ of vision tokens with the highest scores.
\textbf{(2) Dynamic patches} correspond to regions that exhibit significant visual changes. To detect them, we divide the current and previous input images into patches aligned with the vision tokens, compute the cosine similarity between corresponding patches in pixel space, and select the top $\tau_\text{dynamic}\%$ with the lowest similarity values.

\begin{figure}[t]
    \centering
    \includegraphics[width=0.9\linewidth]{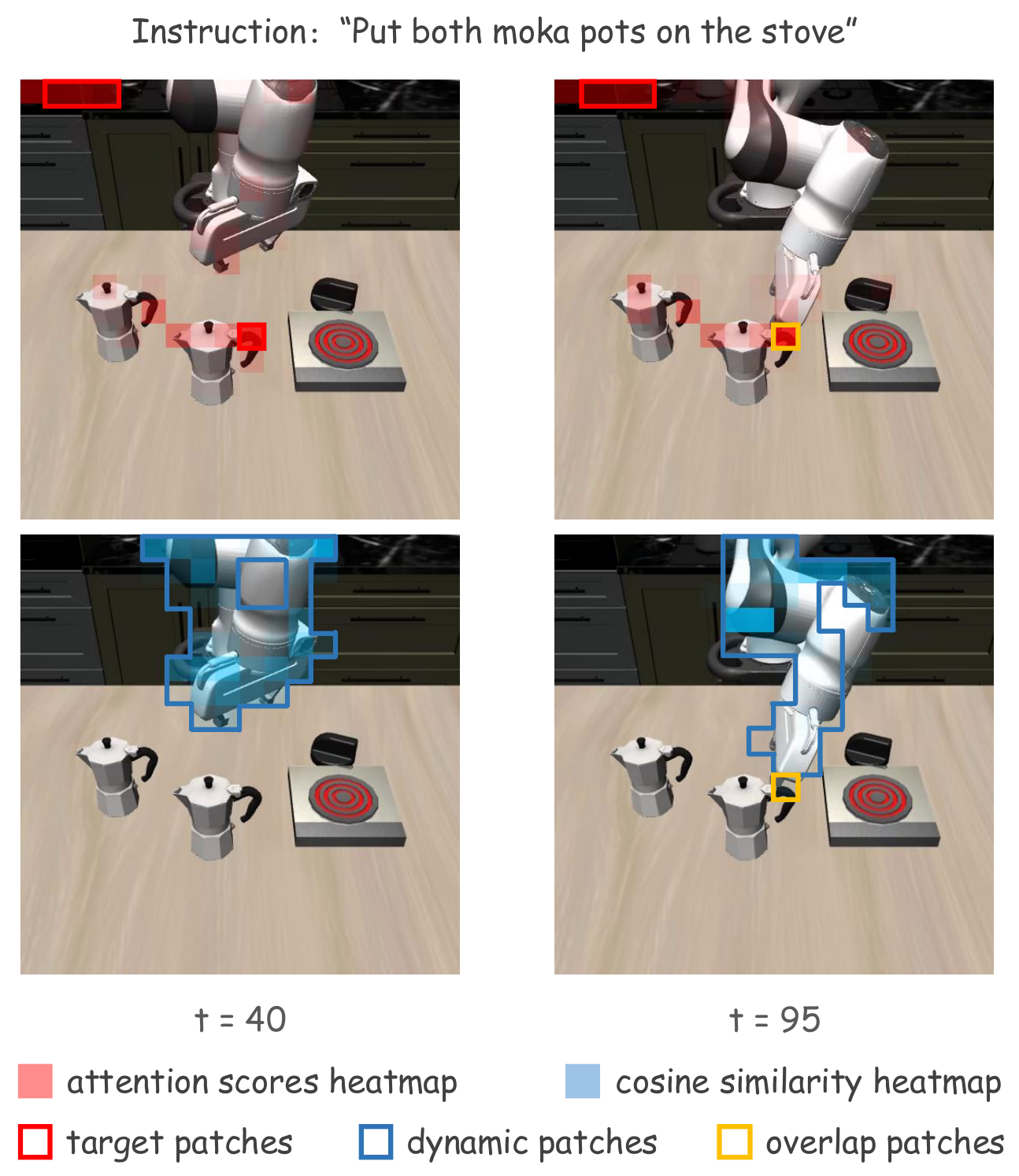}
    \caption{\textbf{Visualization of ATE.}
The attention heatmap uses darker colors to represent higher values, whereas in the cosine similarity heatmap the opposite holds. The parameters are set as $\tau_\text{target} = 1$ and $\tau_\text{dynamic} = 12$.}
    \label{fig:ate}
\end{figure}

Intuitively, dynamic patches capture the motion of the robotic arm and end-effector, while target patches highlight instruction-relevant objects. Overlap between them indicates fine-grained manipulations, such as grasping. As shown in Fig.~\ref{fig:ate}, visualization of both patch types during Mantis inference supports this observation. When such overlap occurs, the Temporal Ensemble is activated to enhance motion stability; otherwise, it is disabled to improve computational efficiency. Integrating ATE into Mantis yields a more efficient variant, termed Mantis-ATE. For more details about ATE, please refer to Appendix~\ref{sec:ate}.

\section{Experiments}
\label{sec:exper}

\begin{table*}[t]
\caption{
\textbf{Comparison on the LIBERO benchmark.} Mantis exhibits superior performance on 3 of 4 tasks and attains the highest average success rate compared to existing baseline methods, demonstrating the effectiveness of leveraging DVF for action prediction. \textbf{Bold} indicates the best performance, and \underline{\textit{Italics}} indicates the second-best performance.
}
\renewcommand{\arraystretch}{1.07}
\setlength{\tabcolsep}{10pt} 

\centering

\label{tab:mainresult}

\begin{tabular}{cc|ccccc}
\toprule[0.5mm]
&  & \textbf{Spatial} & \textbf{Object} & \textbf{Goal} & \multicolumn{1}{c|}{\textbf{Long}} & \makecell[l]{\textbf{Avg.}} \\ 

\midrule
\midrule

\multirow{5}{*}{\makecell{\textit{non-vision-} \\ \textit{augmented}}} 

& Diffusion Policy~\cite{chi2025diffusion} & 78.3    & 92.5   & 68.3 & \multicolumn{1}{c|}{50.5} & 72.4 \\
& OpenVLA~\cite{kim2024openvla}          & 84.7    & 88.4   & 79.2 & \multicolumn{1}{c|}{53.7} & 76.5 \\
& $\pi_0$~\cite{black2024pi_0}       & 96.8    & \underline{\textit{98.8}}   & \textbf{95.8} & \multicolumn{1}{c|}{85.2} & 94.2 \\
& $\pi_0$-FAST~\cite{pertsch2025fast}          & 96.4    & 96.8   & 88.6 & \multicolumn{1}{c|}{60.2} & 85.5 \\
& NORA~\cite{hung2025nora}                 & 92.2    & 95.4   & 89.4 & \multicolumn{1}{c|}{74.6} & 87.9 \\ 

\midrule

\multirow{8}{*}{\makecell{\textit{vision-} \\ \textit{augmented}}} 

& ATM~\cite{wen2023any}              & 68.5    & 68.0   & 77.8 & \multicolumn{1}{c|}{39.3} & 63.4 \\
& CoT-VLA~\cite{zhao2025cot}          & 87.5    & 91.6   & 87.6 & \multicolumn{1}{c|}{69.0} & 81.1 \\
& WorldVLA~\cite{cen2025worldvla}         & 87.6    & 96.2   & 83.4 & \multicolumn{1}{c|}{60.0} & 81.8 \\
& UniVLA~\cite{bu2025learning}           & 96.5    & 96.8   & \underline{\textit{95.6}} & \multicolumn{1}{c|}{92.0} & 95.2 \\
& UnifiedVLA~\cite{wang2025unified}       & 95.4    & \underline{\textit{98.8}}   & 93.6 & \multicolumn{1}{c|}{\underline{\textit{94.0}}} & 95.5 \\
& DreamVLA~\cite{zhang2025dreamvla}         & 97.5    & 94.0   & 89.5 & \multicolumn{1}{c|}{89.5} & 92.6 \\
& $\mathcal{F}_1$~\cite{lv2025f1}  & \underline{\textit{98.2}}    & 97.8   & 95.4 & \multicolumn{1}{c|}{91.3} & \underline{\textit{95.7}} \\
& \cellcolor[RGB]{230,239,230}Mantis (Ours)    & \cellcolor[RGB]{230,239,230}\textbf{98.8}    & \cellcolor[RGB]{230,239,230}\textbf{99.2}   & \cellcolor[RGB]{230,239,230}94.4 & \multicolumn{1}{c|}{\cellcolor[RGB]{230,239,230}\textbf{94.2}} & \cellcolor[RGB]{230,239,230}\textbf{96.7} \\ 

\bottomrule[0.5mm]
\end{tabular}
\end{table*}

\subsection{Implementation Details}


\textbf{Basic configuration.} 
Mantis comprises 5.8 billion parameters in total: 3.7B in the backbone, 1.4B in the DVF head, 0.3B in the action head, and 0.3B in the VAE. We set the number of $\texttt{[LAT]}$ to 9, $\texttt{[ACT]}$ to 6, and $\texttt{[GAP]}$ to $6 \times 3$ corresponding to time step gaps from 1 to 6. The diffusion process uses 30 steps for the DVF head and 10 steps for the action head. 
Training adopts AdamW~\cite{loshchilov2017decoupled} with 0.1 weight decay and 0.5 gradient clipping, and leverages DeepSpeed~\cite{rasley2020deepspeed} for efficient distributed training.

\vspace{0.5em}
\noindent
\textbf{Pretraining Setup.} 
In \textit{Stage 1}, the model is pretrained on the SSV2 dataset~\cite{goyal2017something}, which contains approximately 220K videos of human manipulations. The time step gap is randomly sampled between 1 and 6.
In \textit{Stage 2}, the model is pretrained on the DROID dataset~\cite{khazatsky2024droid}, consisting of 76K robot episodes. The vision loss weight $\alpha$ is set to 0.1. 
In \textit{Stage 3}, we introduce language supervision by jointly training on 38 multimodal datasets and DROID for 1.5 epochs.

\vspace{0.5em}
\noindent
\textbf{Finetuning Setup.}
For downstream finetuning on the LIBERO benchmark~\cite{liu2023libero}, we use the same learning rate configuration as in pretraining stages 1 and 2 and train for 30 epochs without language supervision. The vision loss weight is set to $\alpha = 0.1$. The checkpoint with the highest validation success rate (SR) is selected for final evaluation. Further implementation details are provided in Appendix~\ref{sec:impl}.

\subsection{Simulation Experiments}

\begin{figure}[t]
    \centering
    \includegraphics[width=1.0\linewidth]{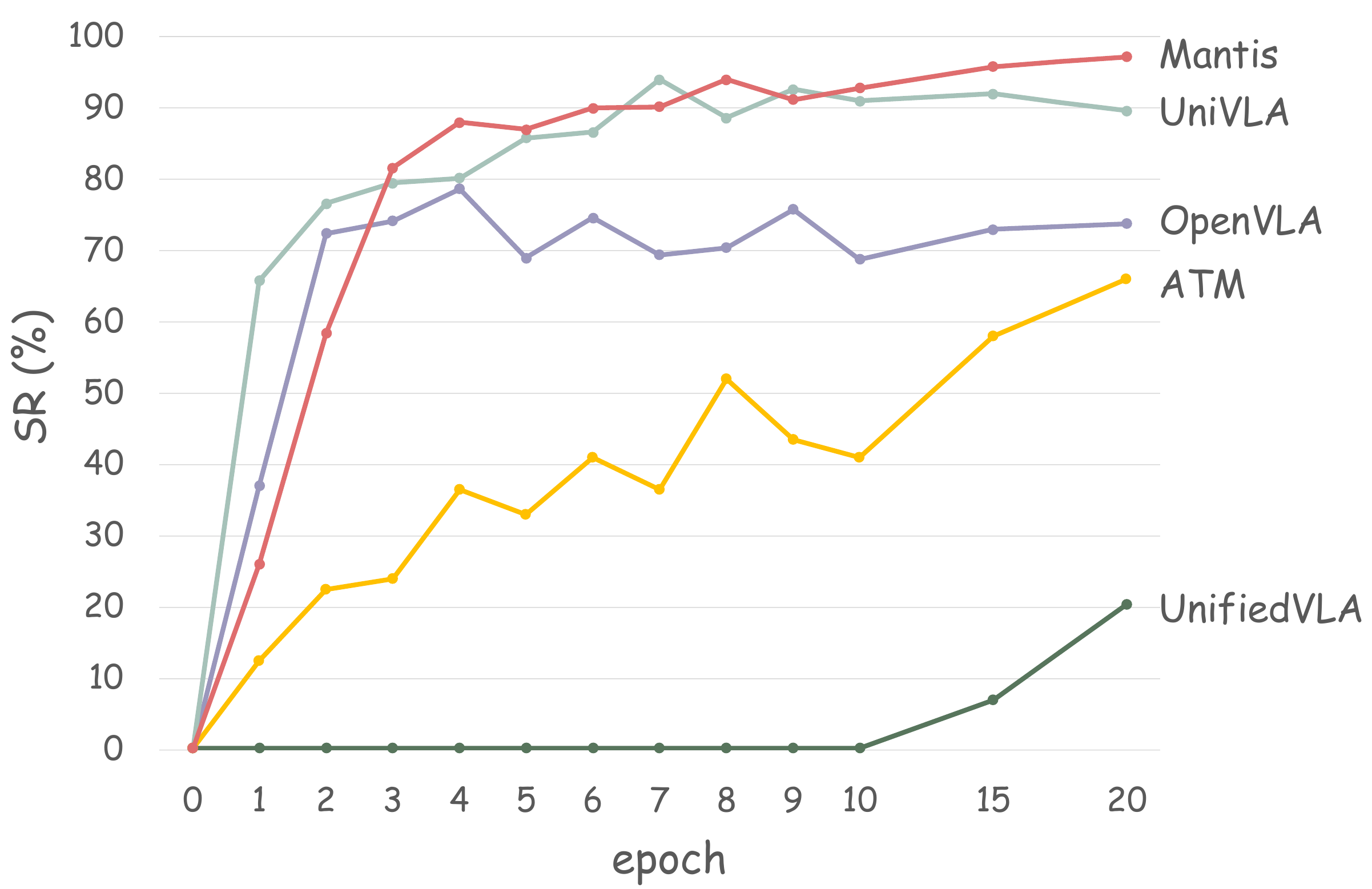}
    \caption{\textbf{Convergence speed comparison.} Compared with traditional visual foresight methods such as UnifiedVLA~\cite{wang2025unified}, Mantis achieves significantly faster convergence speed, underscoring the necessity of decoupling foresight prediction from action learning.
    }
    \label{fig:line_chart}
\end{figure}

\begin{figure*}[t]
    \centering
    \includegraphics[width=0.95\linewidth]{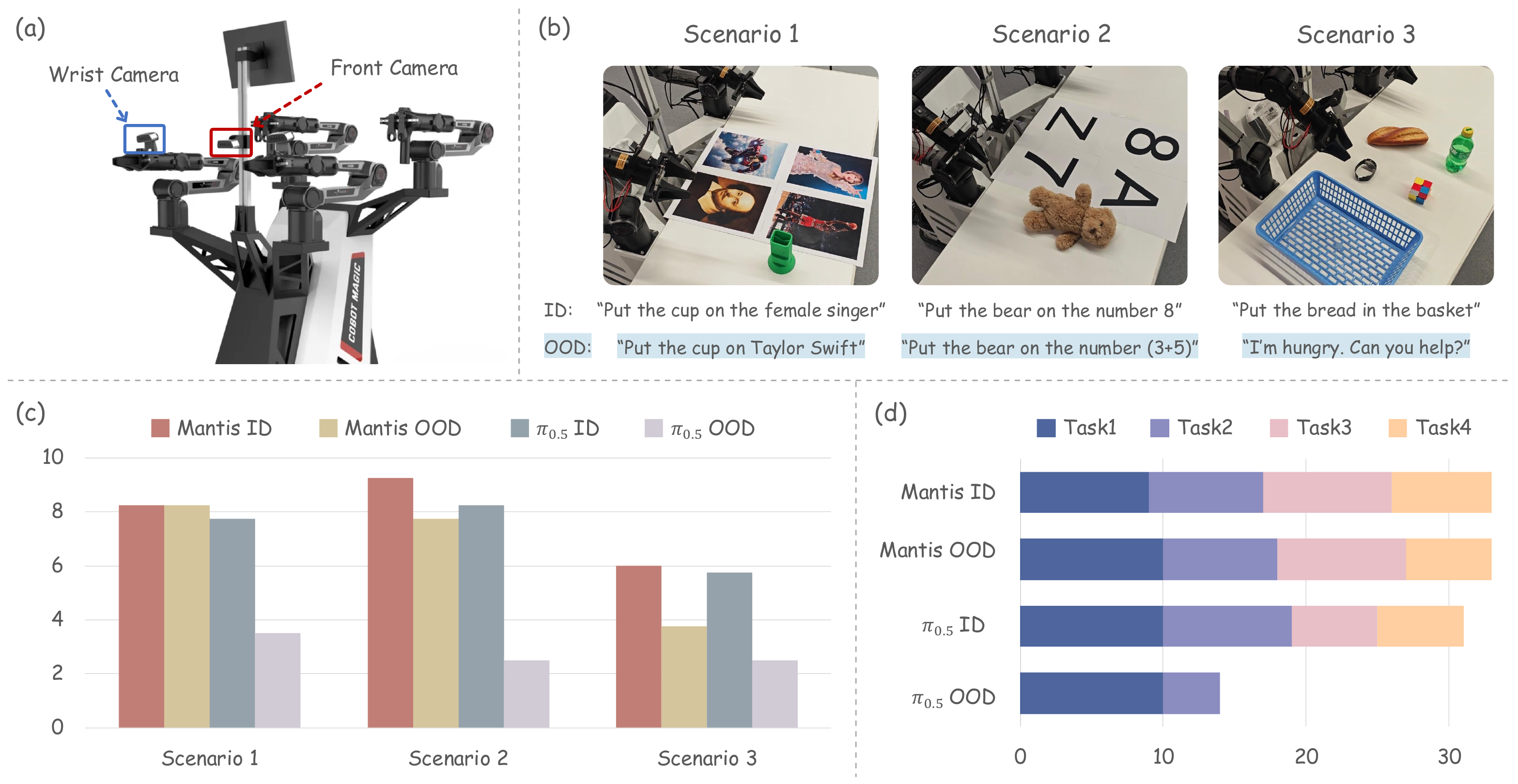}
    \caption{\textbf{Real World Experiments.} (a) The Agilex platform. (b) Scenario setups and example instructions. Each scenario shows one ID instruction and the corresponding OOD instruction. (c) Average success counts for Mantis and $\pi_{0.5}$ on ID and OOD tasks across three scenarios. (d) Per-task success counts for Mantis and $\pi_{0.5}$ in Scenario 1.}
    \label{fig:real_world}
\end{figure*}


We evaluate Mantis on the widely adopted LIBERO benchmark~\cite{liu2023libero}, which comprises four task suites: Spatial, Object, Goal, and Long. 
Each suite contains 10 tasks, with 50 trials per task to reduce random variation. 
Performance is measured using Success Rate (SR) from 0 to 100, where higher is better.
For comprehensive comparison, we select high-performance VLA models both with~\cite{wen2023any, zhao2025cot, cen2025worldvla, bu2025learning, wang2025unified, zhang2025dreamvla, lv2025f1} and without~\cite{chi2025diffusion, kim2024openvla, black2024pi_0, pertsch2025fast, hung2025nora} vision augmentation as baselines. 
We also compare the convergence speed of Mantis against four vision-augmented baselines.

\vspace{0.5em}
\noindent
\textbf{Main Results.} 
As shown in Table~\ref{tab:mainresult}, Mantis exhibits superior performance on 3 of 4 task suites and attains the highest average SR, outperforming previous vision-augmented and non-vision-augmented methods.
This underscores the efficacy of leveraging DVF to enhance action prediction.
Furthermore, vision-augmented methods mostly outperform their non-vision-augmented counterparts, corroborating the findings in~\cite{li2025drivevla} that dense visual states effectively complement sparse action signals.
Notably, ATM~\cite{wen2023any} exhibits inferior performance, which we attribute to the limited accuracy of point trajectories extracted via video tracking methods, resulting in accumulated errors.

\vspace{0.5em}
\noindent
\textbf{Convergence Speed.}
We evaluate the convergence speed of Mantis and four baseline methods representing different learning paradigms: non-vision-augmentation (OpenVLA~\cite{kim2024openvla}), visual foresight (UnifiedVLA~\cite{wang2025unified}), track guidance (ATM~\cite{wen2023any}), and latent action supervision (UniVLA~\cite{bu2025learning}). 
Each model is fine-tuned from its pretrained checkpoint for 20 epochs on the LIBERO Spatial suite following the prescribed configuration. Evaluation is conducted at each epoch, and the results are presented as an epoch-SR line graph in Fig.~\ref{fig:line_chart}.
As illustrated, Mantis demonstrates a relatively fast convergence speed, comparable to the non-vision-augmented OpenVLA and the latent action supervision method UniVLA. 
In contrast, the entangled visual foresight approach, UnifiedVLA, converges the slowest, maintaining a success rate of zero during the first ten epochs. 
This observation highlights the necessity of decoupling foresight prediction from action learning in order to achieve efficient optimization.

\subsection{Real World Experiments}

\begin{figure*}[t]
    \centering
    \includegraphics[width=1.0\linewidth]{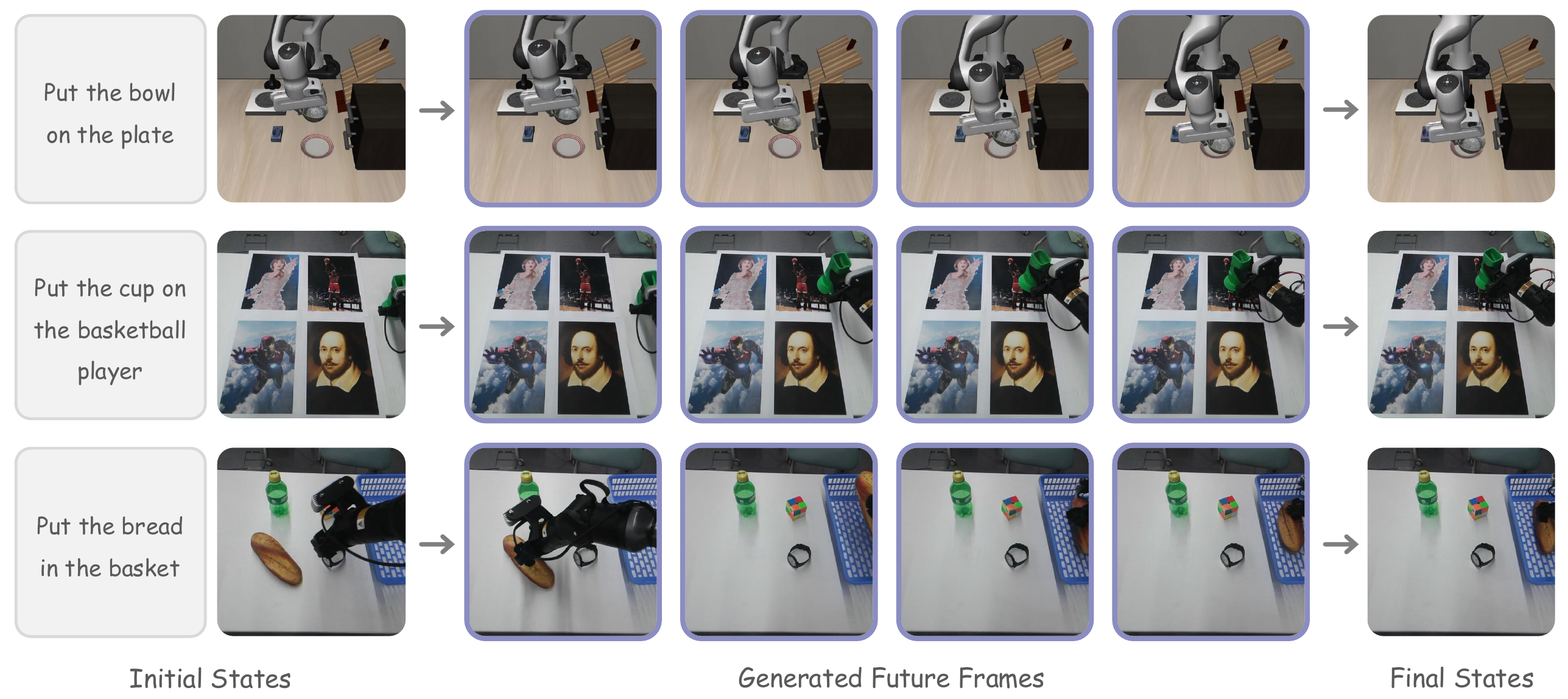}
    \caption{\textbf{Visualization of Generated Future Frames.} The last generated future frame closely mirrors the ground truth final state, substantiating the efficacy of the DVF in refining action prediction across diverse manipulation tasks.}
    \label{fig:samples}
\end{figure*}

\begin{figure}[t]
    \centering
    \includegraphics[width=1.0\linewidth]{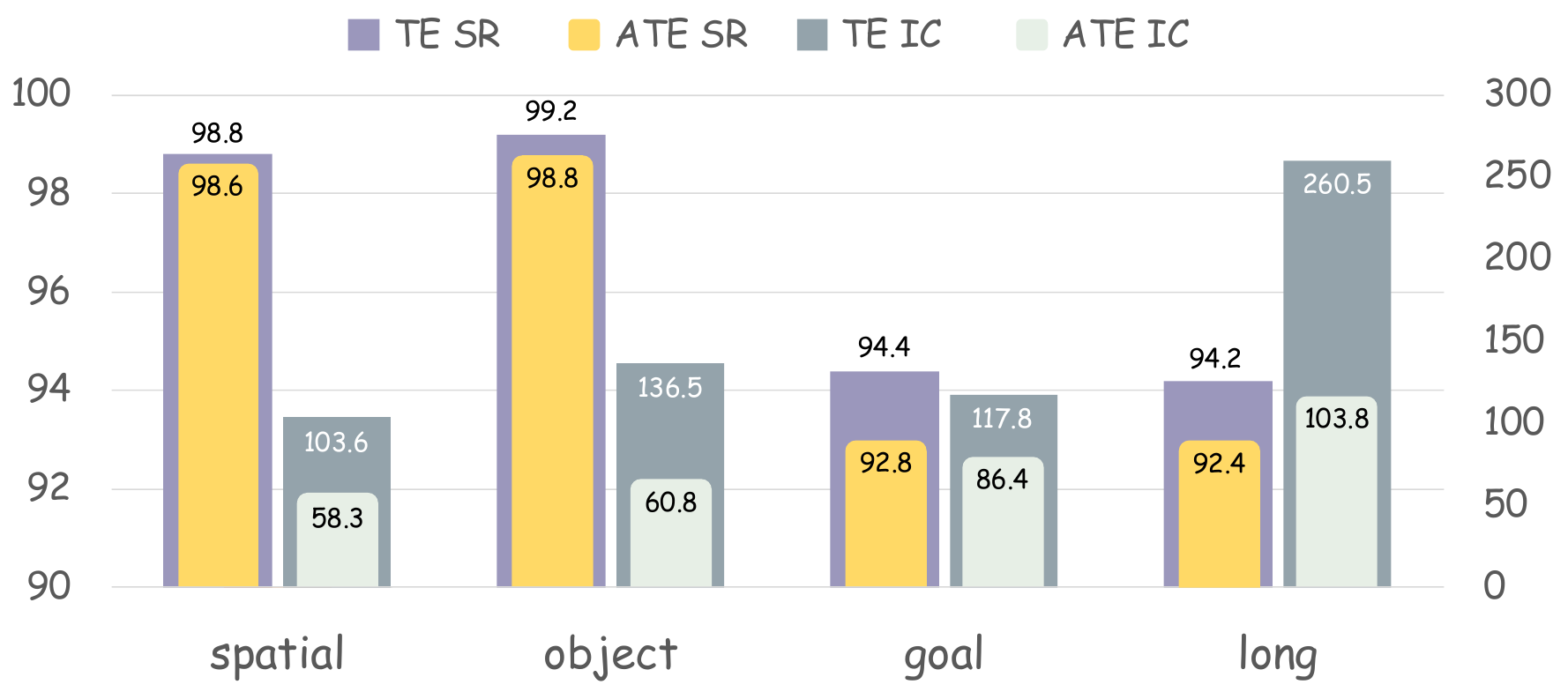}
    \caption{\textbf{Comparison between standard Mantis (TE) and Mantis-ATE.} The primary vertical axis denotes success rate (SR), and the secondary vertical axis denotes inference count (IC).}
    \label{fig:ate_results}
\end{figure}

We evaluate Mantis on an Agilex platform (Fig.~\ref{fig:real_world} (a)) to investigate the effect of language supervision on preserving the backbone's capabilities.
We design three experimental scenarios, each containing four in-domain (ID) instructions to assess instruction following and four out-of-domain (OOD) instructions to evaluate generalization.
Fig.~\ref{fig:real_world} (b) shows the setup of the scenarios and provides example instructions.
Effective generalization demands a nuanced understanding and reasoning. For example, the first scenario requires world knowledge (\eg, Taylor Swift), while the second involves basic arithmetic logic.
We compare Mantis with $\pi_{0.5}$~\cite{intelligence2025pi_}, a state-of-the-art open-source VLA model. Both are fine-tuned separately on the three scenarios, with Mantis maintaining language supervision throughout training.
Each instruction is executed 10 times, and performance is reported as the average number of successful executions, allowing up to 5 consecutive attempts per trial.

As shown in Fig.~\ref{fig:real_world} (c), Mantis consistently outperforms $\pi_{0.5}$ on both ID and OOD instructions across all three scenarios, demonstrating its superior instruction-following  capability, generalization to unseen instructions, and fundamental reasoning ability. 
In contrast, $\pi_{0.5}$ exhibits subpar instruction-following ability and almost no generalization capability to OOD instructions (Fig.~\ref{fig:real_world} (d)).
This corroborates the efficacy of language supervision in preserving the backbone's understanding and reasoning capability.
Additionally, we activated the DVF head in certain experiments to collect the generated future frame sequences, as visualized in Fig.~\ref{fig:samples}.
Further details are provided in Appendix~\ref{sec:real}.


\vspace{-0.5em}
\subsection{Ablation Study}

\textbf{ATE Analysis.}
We evaluated ATE’s impact on task execution speed by comparing the standard Mantis (TE) and Mantis-ATE across four LIBERO suites. 
We measured the average inference count (IC) and success rate (SR), where a lower IC indicates higher efficiency. 
The parameters are set as $\tau_\text{target} = 1$ and $\tau_\text{dynamic} = 12$.
As shown in Fig.~\ref{fig:ate_results}, Mantis-ATE reduces the IC by nearly 50\% while maintaining comparable performance, demonstrating a substantial improvement in inference efficiency.


\vspace{0.5em}
\noindent
\textbf{DVF Ablations.} We evaluated the effect of the DVF through four architectural variants across four LIBERO suites:
(1) vanilla-DVF, 
(2) flawed-DVF (DVF without the residual connection), 
(3) no-DVF (only the action head), and 
(4) pretrained-DVF (DVF pretrained on human and robot videos).
The first three models were trained from scratch, while all four were trained for 30 epochs on each task suite. The best-performing checkpoint from each model was used for evaluation.
The results, summarized in Table~\ref{tab:ablation}, show that the pretrained-DVF achieved the highest success rate (SR), followed by the vanilla-DVF, then the flawed-DVF, with the no-DVF configuration performing the worst.
These results support the following conclusions:
(a) DVF facilitates action learning,
(b) the residual connection enables DVF to better capture latent actions, and
(c) video pretraining further enhances the performance of DVF.

\vspace{0.5em}
\noindent
\textbf{Language Supervision Ablations.} 
We assessed the understanding and reasoning abilities of Mantis using multiple multimodal benchmarks and performed ablation studies without language supervision in real-world settings. The results are provided in Appendix~\ref{sec:lang}.

\begin{table}[t]

\caption{\textbf{Comparison of four DVF variants.} \textbf{Bold} indicates the best performance, \underline{\textit{Italics}} indicates the second-best performance.}

\renewcommand{\arraystretch}{1.07}
\centering
\small

\label{tab:ablation}

\begin{tabular}{c|cccc|c}
\toprule[0.5mm]
& Spatial & Object & Goal & Long & Avg. \\ 

\midrule
\midrule

vanilla-DVF  & \underline{\textit{98.2}}    & \underline{\textit{98.8}}   & 93.6 & \underline{\textit{92.2}} & \underline{\textit{95.7}} \\
flawed-DVF & 96.8    & 97.0   & \underline{\textit{93.8}} & 89.8 & 94.4 \\
no-DVF & 93.4    & 93.0   & 90.4 & 88.2 & 91.3 \\ 
pretrained-DVF  & \textbf{98.4}    & \textbf{99.0}   & \textbf{94.0} & \textbf{93.2} & \textbf{96.2} \\

\bottomrule[0.5mm]
\end{tabular}
\end{table} 

\section{Conclusion, Limitations and Future Work}
\label{sec:conclusion}

In this work, we present Mantis, a framework featuring Disentangled Visual Foresight (DVF). Simulation experiments demonstrate that DVF enhances performance and convergence speed, while real-world experiments validate the effectiveness of language supervision.
Limitations include minor motion rollbacks in real-world scenarios due to missing robot state inputs. Future work will integrate richer inputs (\eg, 3D point clouds) and optimize inference speed.






{
    \small
    \bibliographystyle{ieeenat_fullname}
    \bibliography{main}
}

\appendix
\clearpage
\setcounter{page}{1}

\begin{table}[b]

\caption{
\textbf{Comparison on VQA and multimodal understanding benchmarks.}
Mantis achieves superior performance on 2 of the 3 benchmarks. Compared with the original backbone, its performance decreases only marginally. \textbf{Bold} denotes the best results.
}
\renewcommand{\arraystretch}{1.07}
\setlength{\tabcolsep}{7pt} 

\centering
\small

\label{tab:language}

\begin{tabular}{c|ccc}
\toprule[0.5mm]
             & MME    & OCRBench & RealWorldQA \\ 
\midrule
Qwen2.5-VL   & 2217.3 & 807      & 62.1        \\ 
\midrule
ECoT         & 0      & 12       & 0           \\
ChatVLA      & 1435.2 & 729      & \textbf{57.0}        \\
Mantis(Ours) & \textbf{2070.2} & \textbf{757}      & 56.9        \\ 
\bottomrule[0.5mm]
\end{tabular}
\end{table}

\begin{figure}[b]
    \centering
    \includegraphics[width=0.95\linewidth]{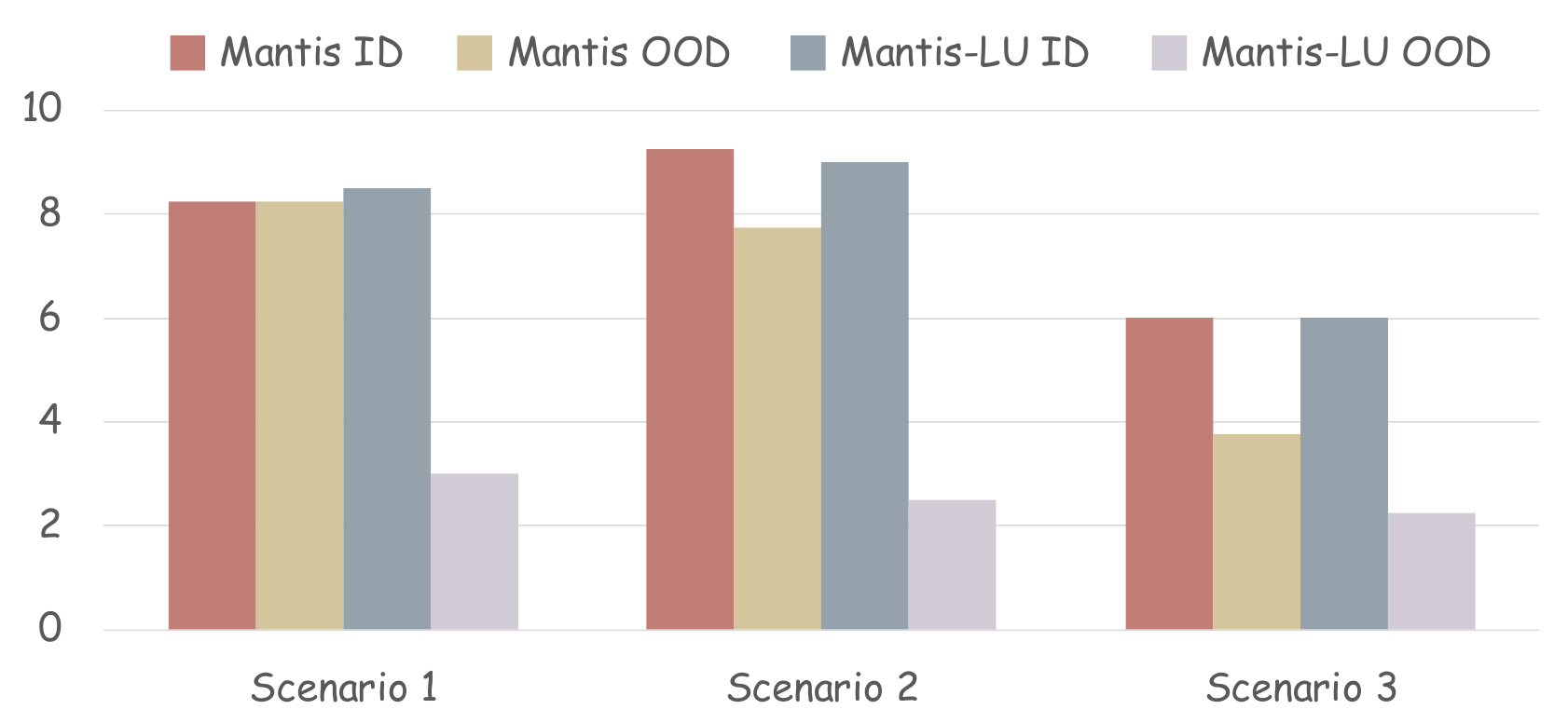}
    \caption{\textbf{Comparison between Mantis and Mantis-LU.}}
    \label{fig:real_world_language}
\end{figure}

\section{Adaptive Temporal Ensemble}
\label{sec:ate}


When applying ATE, each input image is divided into $18 \times 18$ patches. Unless otherwise specified, we set $\tau_{\text{target}} = 1$ and $\tau_{\text{dynamic}} = 12$ for all experiments. We then analyze the computational complexity. For standard Mantis inference, the theoretical FLOPs for each backbone layer are:
\begin{equation}
  \text{FLOPs} \approx 4 L D^2 + 2 L^2 D + 2 L D M
  \label{eq:flops}
\end{equation}
where $L$ is the number of input tokens to the first backbone layer, $D$ is the hidden-state dimension, and $M$ denotes the feed-forward network (FFN) intermediate dimension.
Dynamic-patch identification introduces an additional cost of $\mathcal{O}(L_v D_{\text{patch}})$, incurred by patch-similarity computations, where $L_v$ is the number of image patches and $D_{\text{patch}}$ is the dimension of each patch. 
Target-patch selection further requires cross-modal attention aggregation, adding $\mathcal{O}(L_t L_v D)$, with $L_t$ representing the number of text tokens. The thresholding step includes a sorting operation of complexity $\mathcal{O}(L_v \log L_v)$. Empirically, ATE reduces the number of inferences by more than 40\%, and the overhead from dynamic- and target-patch selection is negligible relative to the computation in Eq.~\ref{eq:flops}. As a result, ATE delivers an effective acceleration of the inference process.

\section{Implementation Details}
\label{sec:impl}

For visual inputs, video frames and the robot's primary camera images are cropped and resized to $512 \times 512$ pixels, while the robot's wrist camera images are at $256 \times 256$.
Both training stage 1 and 2 are trained for one epoch using a cosine learning rate schedule with 500 warm-up steps, a base learning rate of $1e-4$, and a minimum learning rate of $1e-5$. 
For stage 3, a fixed learning rate of $1e-5$ is used, with vision loss weight $\alpha = 0.1$ and language loss weight $\beta = 0.005$.
Our 38 language-supervision datasets are sourced from those used in the instruction-training stage of LLaVA-OneVision-1.5-Instruct~\cite{an2025llava}. We include datasets covering visual question answering, OCR, embodied planning, and other general-purpose tasks, while excluding those focused on chart question answering, medical imaging, and other highly specialized domains. A complete list of the selected datasets is provided in Table~\ref{tab:datasets}.

\section{Real World Experiments}
\label{sec:real}

\begin{figure*}[b]
    \centering
    \includegraphics[width=0.98\linewidth]{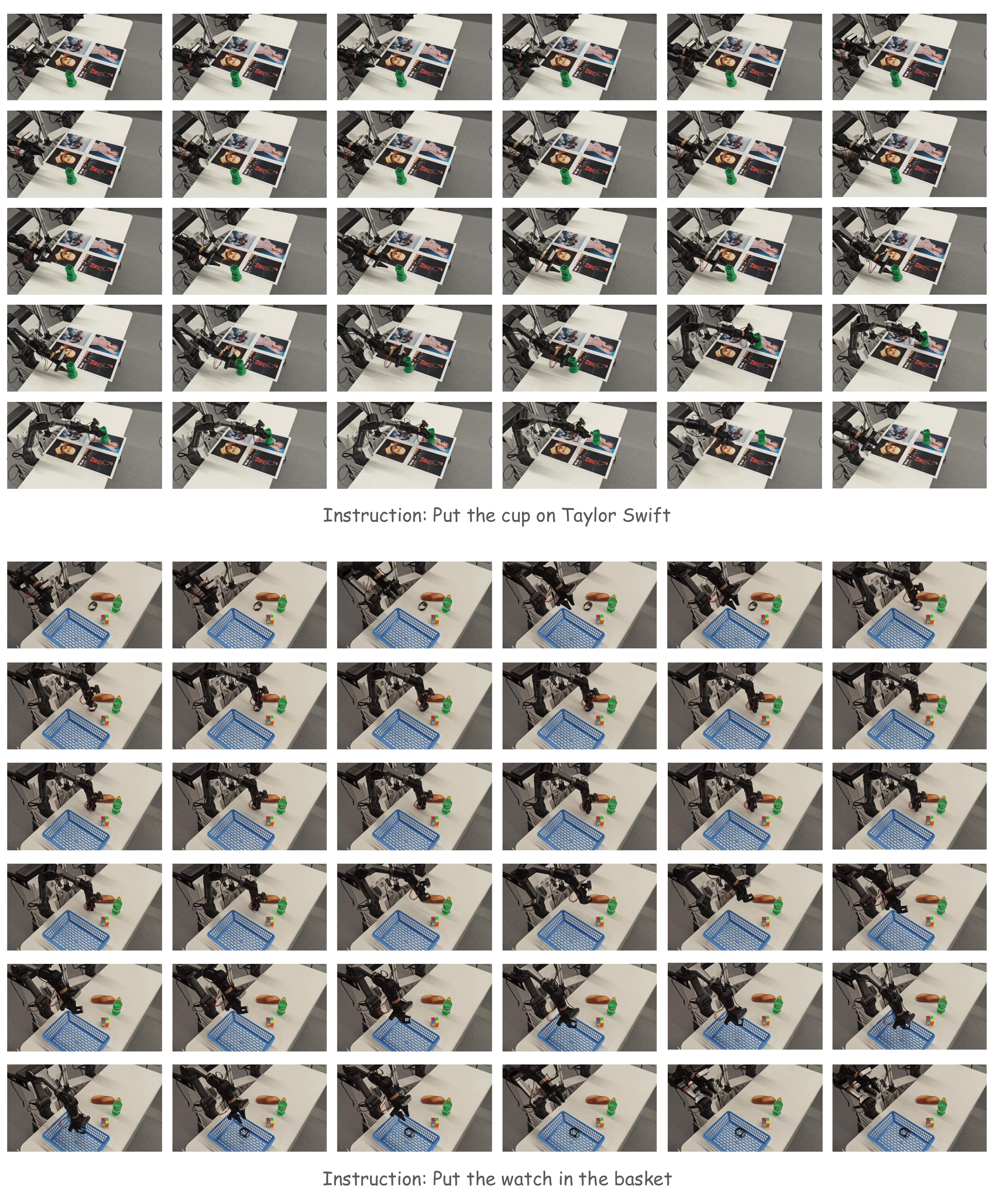}
    \caption{\textbf{Execution examples on real-world tasks.}}
    \label{fig:real_world_execution_examples}
\end{figure*}

All in-domain (ID) and out-of-domain (OOD) instructions for the three scenarios are summarized in Table~\ref{tab:real_world}.
For each scenario, we collected 100 teleoperated demonstration episodes, with 25 episodes per task.
Both Mantis and $\pi_{0.5}$ were fine-tuned on the combined dataset containing all four tasks within each scenario.
For each model, we adopted the same learning-rate settings used in pretraining stage 1 and trained for 10 epochs. Mantis was supervised with the LLaVA-Instruct dataset \cite{liu2023visual}, whereas $\pi_{0.5}$ did not use any language supervision.
We set the vision loss weight to $\alpha = 0.1$ and the language loss weight for Mantis to $\beta = 0.1$.
Task-level success counts for both models across the three scenarios are reported in Table~\ref{tab:real_world_results}.
Execution examples on real-world tasks is shown in Fig.~\ref{fig:real_world_execution_examples}.

\section{Language Supervision Ablations}
\label{sec:lang}
To assess whether language supervision preserves the backbone’s capabilities, we conducted ablation studies comparing Mantis with other VLA models~\cite{zhou2025chatvla, zawalski2024robotic} on several VQA and multimodal understanding benchmarks~\cite{yin2024survey, liu2024ocrbench}.
The original Qwen2.5-VL model was also included for reference, and the results are shown in Table~\ref{tab:language}. Mantis achieves the best performance on two of the three benchmarks, with only a marginal drop relative to the original backbone, confirming the effectiveness of the language supervision.

In the Real-World Experiments, we also trained a language-unsupervised variant, Mantis-LU. Figure~\ref{fig:real_world_language} compares its performance with the original Mantis on both in-domain (ID) and out-of-domain (OOD) instructions across the three scenarios. While Mantis-LU retains reasonably strong performance on ID instructions, it performs poorly on OOD instructions, indicating that language supervision is crucial for instruction generalization.

\begin{table*}[t]

\caption{
\textbf{The 38 datasets used for Mantis language supervision.} These datasets are sourced from the instruction-tuning data of LLaVA-OneVision-1.5-Instruct~\cite{an2025llava} and cover general visual-language tasks while excluding specialized domains.}
\renewcommand{\arraystretch}{1.2}
\setlength{\tabcolsep}{7pt} 
\centering
\small
\label{tab:datasets}

\begin{tabular}{ccccc}
\toprule[0.5mm]
\midrule
alfworld           & allava\_instruct\_laion4v & allava\_instruct\_vflan4v & allava             & cambrian           \\
coco               & Evol-Instruct-GPT4-Turbo  & gpt4o                     & gpt4v              & gqa                \\
laion\_220k        & infographic\_azuregpt4v   & llava\_cot\_100k          & llava\_instruct    & llava\_wild        \\
llrv\_gpt4v        & magpie\_pro               & magpie\_ultra             & magpie\_ultra      & open\_orca         \\
orca\_994k         & orca\_agentinstruct       & sharegpt4o                & sherlock           & vflan              \\
CLEVR-Math         & image\_textualization     & Super-CLEVR               & wikipedia\_2m      & textocr\_gpt4v     \\
sharegpt4v-part-00 & sharegpt4v-part-01        & sharegpt4v-part-03        & sharegpt4v-part-04 & sharegpt4v-part-05 \\
sharegpt4v-part-06 & sharegpt4v-part-07        & sharegpt4v-part-08        &                    &                    \\ 
\midrule
\bottomrule[0.5mm]
\end{tabular}
\end{table*}

\begin{table*}[b]

\caption{
\textbf{In-distribution (ID) and out-of-distribution (OOD) instructions for the three scenarios.} Each scenario comprises four ID instructions and four corresponding OOD instructions. The OOD instructions evaluate world knowledge, basic reasoning abilities, and understanding of human intent across the three scenarios, respectively.
}
\renewcommand{\arraystretch}{1.2}
\setlength{\tabcolsep}{10pt} 
\centering
\label{tab:real_world}

\begin{tabular}{clll}
\toprule[0.5mm]
\multicolumn{2}{l|}{}                                    & \multicolumn{1}{c|}{ID instructions}  & \multicolumn{1}{c}{OOD instructions}             \\ \midrule \midrule
\multicolumn{1}{c|}{\multirow{4}{*}{Scenario 1}} & \multicolumn{1}{l|}{Task1} & \multicolumn{1}{l|}{Put the cup on the female singer}      & Put the cup on Taylor Swift                      \\
\multicolumn{1}{c|}{}                            & \multicolumn{1}{l|}{Task2} & \multicolumn{1}{l|}{Put the cup on the basketball player}  & Put the cup on Michael Jordan                    \\
\multicolumn{1}{c|}{}                            & \multicolumn{1}{l|}{Task3} & \multicolumn{1}{l|}{Put the cup on the Marvel superhero}   & Put the cup on Iron Man                          \\
\multicolumn{1}{c|}{}                            & \multicolumn{1}{l|}{Task4} & \multicolumn{1}{l|}{Put the cup on the English playwright} & Put the cup on Shakespeare                       \\ \midrule
\multicolumn{1}{c|}{\multirow{4}{*}{Scenario 2}} & \multicolumn{1}{l|}{Task1} & \multicolumn{1}{l|}{Put the bear on the number 8}          & Put the bear on the number (3+5)                 \\
\multicolumn{1}{c|}{}                            & \multicolumn{1}{l|}{Task2} & \multicolumn{1}{l|}{Put the bear on the letter A}          & Put the bear on the first letter                 \\
\multicolumn{1}{c|}{}                            & \multicolumn{1}{l|}{Task3} & \multicolumn{1}{l|}{Put the bear on the letter Z}          & Put the bear on the last letter                  \\
\multicolumn{1}{c|}{}                            & \multicolumn{1}{l|}{Task4} & \multicolumn{1}{l|}{Put the bear on the number 7}          & Put the bear on the number (9-2)                 \\ \midrule
\multicolumn{1}{c|}{\multirow{4}{*}{Scenario 3}} & \multicolumn{1}{l|}{Task1} & \multicolumn{1}{l|}{Put the bottle in the basket}          & Put a thing that can quench thirst in the basket \\
\multicolumn{1}{c|}{}                            & \multicolumn{1}{l|}{Task2} & \multicolumn{1}{l|}{Put the Rubik's Cube in the basket}    & I want to play with something. Can you help?     \\
\multicolumn{1}{c|}{}                            & \multicolumn{1}{l|}{Task3} & \multicolumn{1}{l|}{Put the bread in the basket}           & I'm hungry. Can you help?                        \\
\multicolumn{1}{c|}{}                            & \multicolumn{1}{l|}{Task4} & \multicolumn{1}{l|}{Put the watch in the basket}                                & Put a thing that can tell the time in the basket \\ \bottomrule[0.5mm]
\end{tabular}
\end{table*}

\begin{table*}[t]

\caption{
\textbf{Task-level success counts.} Mantis outperforms $\pi_{0.5}$ on both in-domain (ID) and out-of-domain (OOD) instructions, demonstrating strong instruction-following and generalization capabilities through effective language supervision.
}
\renewcommand{\arraystretch}{1.1}
\setlength{\tabcolsep}{9pt} 
\label{tab:real_world_results}
\centering

\begin{tabular}{cc|ccccc|ccccc|ccccc}
\toprule[0.5mm]
\multicolumn{2}{c|}{Scenario}                                   & \multicolumn{5}{c|}{\textit{1}} & \multicolumn{5}{c|}{\textit{2}} & \multicolumn{5}{c}{\textit{3}} \\ \midrule
\multicolumn{2}{c|}{Task}                                       & \textit{1}   & \textit{2} & \textit{3} & \textit{4} & \textit{Avg.} & \textit{1}  & \textit{2} & \textit{3} & \textit{4} & \textit{Avg.} & \textit{1}  & \textit{2} & \textit{3} & \textit{4} & \textit{Avg.} \\ 

\midrule
\midrule

\multicolumn{1}{c|}{\multirow{2}{*}{Mantis}} & ID  & 9   & 8 & 9 & 7 & 8.25 & 9  & 10 & 9 & 9 & 9.25 & 6  & 7 & 6 & 5 & 6    \\
\multicolumn{1}{c|}{}                        & OOD & 10  & 8 & 9 & 6 & 8.25 & 9  & 8  & 9 & 6 & 7.75 & 6  & 0 & 6 & 3 & 3.75 \\ \midrule
\multicolumn{1}{c|}{\multirow{2}{*}{$\pi_{0.5}$}}     & ID  & 10  & 9 & 6 & 6 & 7.75 & 10 & 6  & 8 & 9 & 8.25 & 5  & 7 & 5 & 6 & 5.75 \\
\multicolumn{1}{c|}{}                        & OOD & 10  & 4 & 0 & 0 & 3.5  & 10 & 0  & 0 & 0 & 2.5  & 0 & 10 & 0 & 0 & 2.5  \\ 
\bottomrule[0.5mm]
\end{tabular}
\end{table*}

\end{document}